\documentclass[11pt]{article}
\usepackage[table]{xcolor}
\definecolor{grey}{rgb}{0.5,0.5,0.5}
\usepackage{amsfonts}
\usepackage{amsmath}
 \usepackage{multirow}
 \usepackage{algorithm}
\usepackage{algorithmic}
\usepackage{subcaption}
\usepackage{float}
\usepackage{tabularx}
\usepackage[preprint]{acl}

\usepackage{times}
\usepackage{latexsym}
\usepackage{booktabs}
\usepackage[T1]{fontenc}

\usepackage[utf8]{inputenc}

\usepackage{microtype}

\usepackage{inconsolata}

\usepackage{graphicx}

\title{Multi-level context Modeling for consistent expert selection  in Mixture-of-Experts}

\author{
Shuhan Huang$^{1}$ \quad Yuanbo Tang$^{1}$ \quad Naifan Zhang$^{1}$ \quad Yang Li$^{2}$ \quad Wai Kin Victor Chan$^{1}$ \\
$^{1}$Tsinghua Shenzhen International Graduate School, Tsinghua University \\
$^{2}$School of AI, The Chinese University of Hong Kong (Shenzhen)
}

\begin{document}
\maketitle
\begin{abstract}

Mixture-of-Experts (MoE) enables efficient scaling of Transformer models by routing tokens to a small subset of experts. However, existing routers typically condition expert selection on shallow or isolated token representations, which often produce unstable and semantically inconsistent routing decisions across layers. In this work, we revisit expert selection from a representation perspective and identify context incompleteness as a key bottleneck limiting effective expert specialization. To address this issue, we propose \textbf{Multi-level Context Fusion MOE (MCF-MOE)}, a framework that constructs context-aware representations by integrating complementary signals from cross-layer semantic aggregation and local token-level interactions,  enabling more informative and consistent expert selection. Experiments on language modeling and understanding benchmarks demonstrate that MCF-MOE consistently improves routing consistency and downstream performance over strong MoE baselines, highlighting the importance of contextual completeness in expert routing.
The code is available at \url{https://github.com/shuhanhuang/MCF-MOE}.
\end{abstract}


\section{Introduction}


Scaling dense Transformer models quickly leads to prohibitive training and inference costs \cite{DBLP:conf/emnlp/Wang0L0ZW24}. 
Mixture-of-Experts (MoE) alleviates this issue by activating only a subset of experts per token, enabling much larger parameter capacity and often outperforming dense models under similar compute budgets \cite{riquelme2021scaling,csordas2024moeut,zhang2024mpmoe}.

\begin{figure}[h]
\centering
\includegraphics[width=0.5\textwidth]{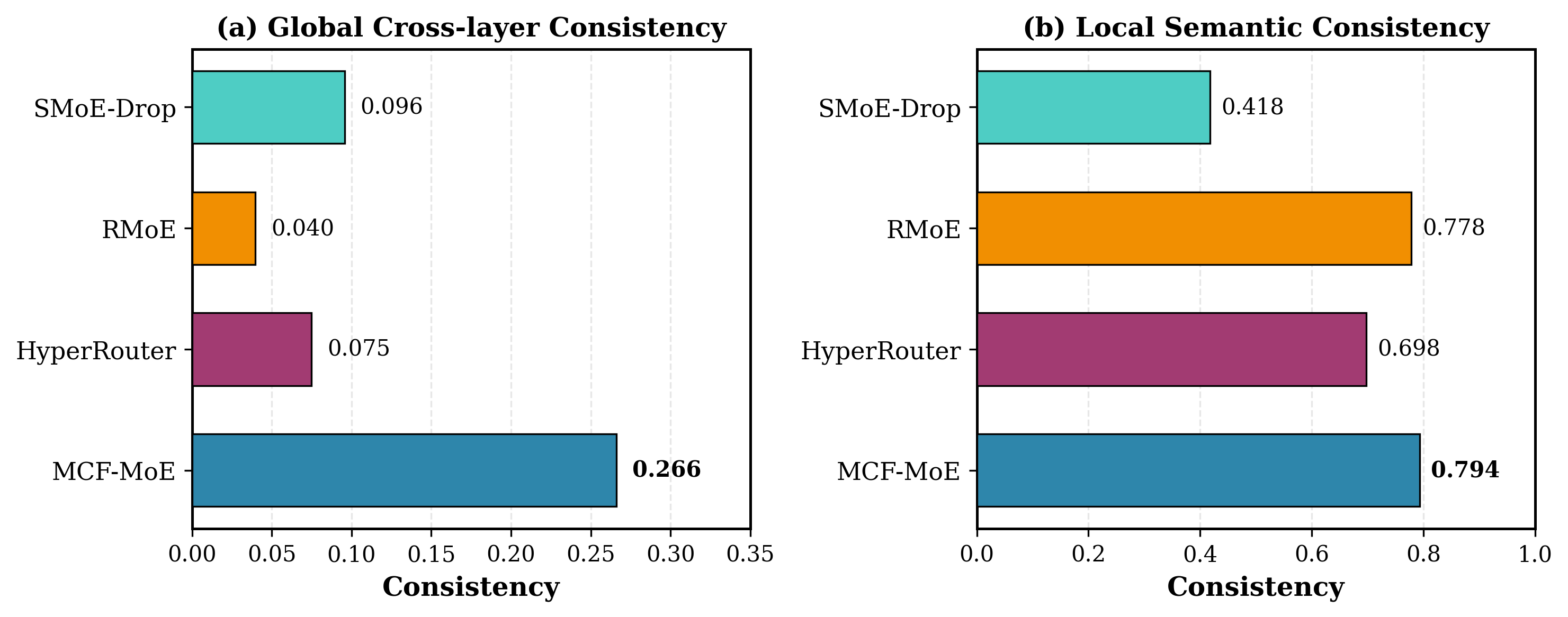} 
\caption{Routing consistency analysis across representative MoE routers.
Existing routers exhibit low cross-layer consistency and unstable semantic routing behavior. 
In contrast, MCF-MOE significantly improves both metrics, yielding more stable and semantically coherent expert assignments.}               
\label{Fig:illustration}
\end{figure}

The effectiveness of MoE models largely depends on the router, which determines how tokens are assigned to experts \cite{vats2024evolution,xian_revisit,carlos_scale}. Consequently, improving routing strategies has become a major research focus in recent years \cite{DBLP:conf/iclr/YueGCGH025,DBLP:conf/asplos/PanLZSTWL025}. 



Sparse Mixture-of-Experts (SMOE) \cite{DBLP:conf/iclr/ShazeerMMDLHD17,DBLP:journals/jmlr/FedusZS22} has become a widely adopted routing paradigm that activates only a small subset of experts for each token. However, excessive sparsity can lead to representation collapse, where expert outputs become overly similar or dominated by a few experts \cite{DBLP:conf/nips/Chi0HDMPSBSMHW22,DBLP:conf/naacl/DoLT25}. To address this issue, several routing strategies have been proposed. For example, SMOE-Dropout \cite{DBLP:conf/iclr/ChenZJLW23} employs randomly initialized routers with dynamic expert activation. HyperRouter \cite{DBLP:conf/emnlp/DoKPNDNLR0H23} generates routing parameters via a fixed hypernetwork. RMOE \cite{DBLP:conf/iclr/QiuHCZWTF25} introduces GRU-based memory to model expert activations across layers.

\begin{figure*}[t]
\centering
\includegraphics[width=0.9\textwidth]{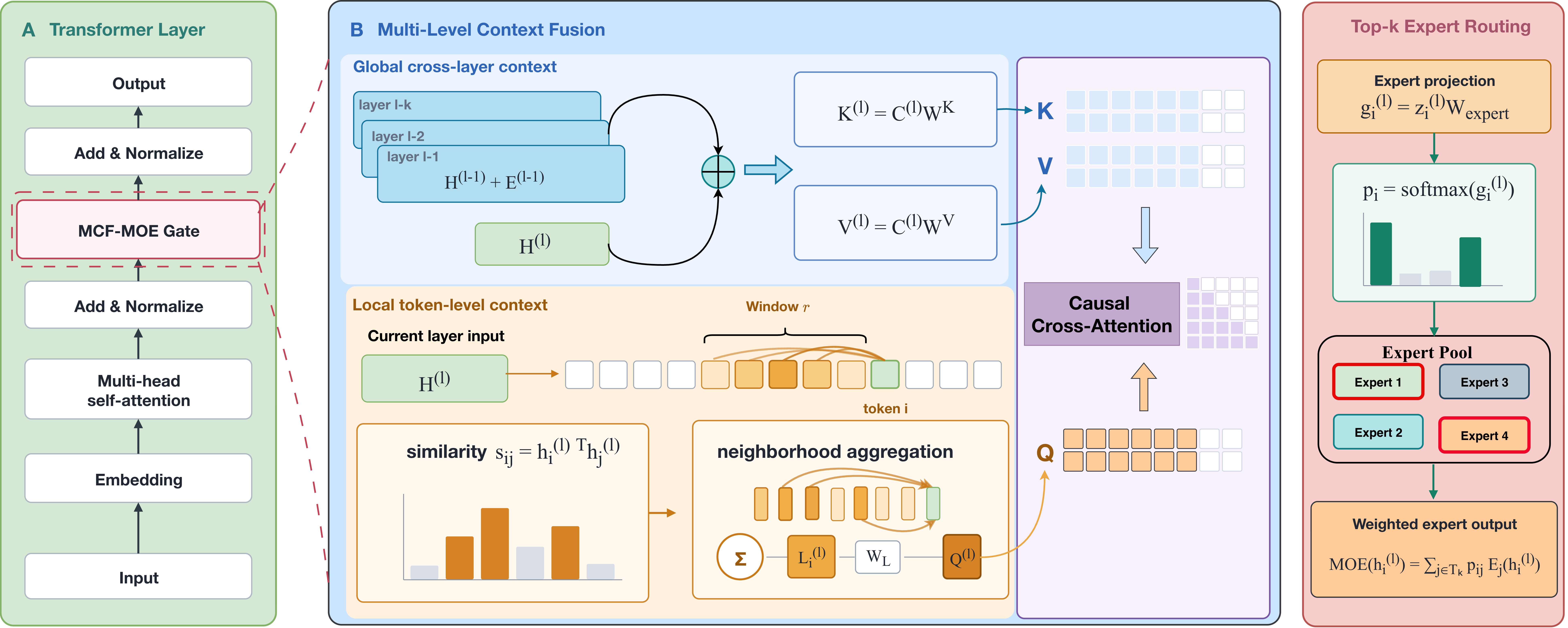} 
\caption{Overall workflow of MCF-MOE. Left: a Transformer layer with the MCF-MOE gate. Middle: local similarity-aware query and global cross-layer keys/values. Right: Top-$K$ expert routing.}
\label{workflow}
\end{figure*}

Recent studies have highlighted the importance of maintaining \textbf{routing consistency} and effectively leveraging contextual information in expert selection. Li et al.~\cite{li2024expert} demonstrate that stronger routing consistency significantly improves both training stability and expert utilization.  Meanwhile, Arnold et al. \cite{DBLP:journals/corr/abs-2409-14107} reveal that expert selection is highly sensitive to contextual semantics, highlighting the need for richer context-aware routing signals.

Despite these advances, existing MoE models still struggle to maintain consistent and semantically coherent expert specialization.  As illustrated in Figure~\ref{Fig:illustration}, we measure both {global cross-layer consistency} and {local semantic consistency} of expert assignments. The results reveal two key observations. First, \textbf{expert assignments exhibit extremely low cross-layer consistency}, suggesting that the underlying representations lack sufficient contextual completeness and fail to capture semantic information consistently across layers. This leads to fragmented expert utilization and unstable specialization patterns.  Second, \textbf{semantic coherence within sequences remains limited and inconsistent} across architectures, indicating that token representations often fail to encode fine-grained contextual relationships necessary for coherent expert behavior.
These observations suggest that the root cause lies not merely in the selection mechanism itself, but more fundamentally in \textbf{under-specified contextual representations}.

To address these limitations, we propose \textbf{Multi-level Context Fusion MoE (MCF-MOE)}, a framework that explicitly models expert selection based on {multi-level contextual evidence}. Figure \ref{workflow} shows the overall workflow of MCF-MOE. Unlike prior works that focus on improving decision mechanisms (e.g., through uncertainty modeling), our approach addresses a fundamentally different problem: how to construct contextually complete representations that better support consistent and semantically aligned expert specialization. Specifically, MCF-MOE integrates two complementary sources of contextual information: (1) \textbf{global cross-layer semantic alignment}, which aggregates contextual signals across layers to capture high-level semantic consistency, and (2) \textbf{local semantic consistency}, which models fine-grained contextual relationships among tokens within a sequence. By jointly modeling these two aspects, MCF-MOE produces more informative and coherent representations, leading to improved consistency across layers and stronger semantic alignment within sequences, thereby facilitating more effective expert collaboration and specialization.

Our main contributions are summarized as follows:
\begin{itemize}
\item We conduct a systematic analysis of expert assignment behavior and identify \textbf{context incompleteness} as a key bottleneck limiting effective expert specialization in MoE architectures.

\item We propose \textbf{MCF-MOE}, a framework that explicitly models multi-level contextual representations by integrating both global cross-layer semantic alignment and local contextual relationships.

\item Extensive experiments demonstrate that improved context modeling leads to more consistent expert specialization and better downstream performance across multiple benchmarks.
\end{itemize}

\section{Related Work}
\subsection{Mitigating Expert Representation Collapse}
To mitigate the long-standing issue of representation collapse in  SMOE \cite{DBLP:conf/iclr/ShazeerMMDLHD17,DBLP:journals/jmlr/FedusZS22} architectures, recent studies address it by enforcing diversity in expert activation. SMOE-Dropout \cite{DBLP:conf/iclr/ChenZJLW23} introduces stochastic gating with frozen routers to balance utilization. X-MoE \cite{DBLP:conf/nips/Chi0HDMPSBSMHW22} imposes geometric constraints via hyperspherical projection, while SimSMOE \cite{DBLP:conf/naacl/DoLT25} penalizes redundancy through cosine and CKA regularization. Stable-MoE \cite{DBLP:conf/acl/Dai0MZSCW22} improves routing stability by freezing the router after a balanced warm-up, and HyperRouter \cite{DBLP:conf/emnlp/DoKPNDNLR0H23} leverages hypernetworks for input-adaptive routing. Collectively, these methods mitigate collapse through stochastic gating, geometric dispersion, redundancy penalization, or dynamic routing, thereby enhancing specialization, robustness, and generalization in SMOE models.
\subsection{Improving Routing Stability and Context Awareness}
As MoE models scale, conventional sparse routing often yields unstable assignments and weak semantic alignment, limiting performance. Recent work addresses these challenges by enhancing stability and context sensitivity. Omi et al. \cite{DBLP:journals/corr/abs-2506-14038} leverages token similarity for relational load balancing. RMOE \cite{DBLP:conf/iclr/QiuHCZWTF25} employs GRU-based memory to enforce cross-layer coherence, while \cite{DBLP:journals/corr/abs-2505-16056} and \cite{DBLP:journals/corr/abs-2409-14107} highlight the roles of local consistency and contextual cues in expert selection.  Taken together, these studies underscore the need for routing strategies that jointly promote stability and semantic alignment in MoE models.

\section{Method}

\subsection{Local Similarity-aware Context Fusion}
\label{sec:local}
To further enhance the model's perception of internal sequence structure, we introduce a local similarity-aware context fusion strategy to model fine-grained semantic associations between tokens within the input sequence. Specifically, we construct a local neighborhood set for each token from the $r$ preceding positions:
 \begin{equation}
 \mathcal{N}(i)=\{j\mid \max(0,\,i-r)\leq j\leq i-1\}
 \end{equation}
where $r$ is the local radius parameter controlling the size of the neighborhood window. The neighborhood is one-sided and excludes the query token itself. We restrict it to $j<i$ to preserve autoregressive consistency, preventing future information from affecting routing. We also exclude $j=i$ to avoid self-similarity dominating the neighborhood selection.
We compute the token-to-token similarity matrix using dot-product attention across the entire sequence:
 \begin{equation}
 \mathbf{S}=\mathbf{H}^{(l)}(\mathbf{H}^{(l)})^\top \in \mathbb{R}^{n \times n}
 \end{equation}
where $\mathbf{S}[i,j]$ represents the similarity score between token $i$ and token $j$.
 A causal local mask is applied to restrict each row to its neighborhood, after which the $k_{\text{loc}}$ most similar tokens are selected for weighted aggregation:
 \begin{equation}
 \mathbf{S}_{\text{masked}}[i,j] = \begin{cases} 
 \mathbf{S}[i,j] & \text{if } j \in \mathcal{N}(i) \\
 -\infty & \text{otherwise}
 \end{cases}
 \end{equation}
 \begin{equation}
 \mathcal{T}(i) = \text{Top-}k_{\text{loc}}(\mathbf{S}_{\text{masked}}[i,:])
 \end{equation}
where $k_{\text{loc}}$ is the number of most similar neighbors to select for each position, and $|\mathcal{T}(i)| = \min(k_{\text{loc}},|\mathcal{N}(i)|)$ so that positions near the start of the sequence simply retrieve fewer neighbors. The locally enhanced representation is then computed as follows:
 \begin{equation}
 \alpha_{ij} = \frac{\exp(\mathbf{S}_{\text{masked}}[i,j])}{\sum_{j' \in \mathcal{T}(i)}\exp(\mathbf{S}_{\text{masked}}[i,j'])}, \quad j \in \mathcal{T}(i)
 \end{equation}
 
  \begin{equation}
 \tilde{\mathbf{h}}_i = \sum_{j \in \mathcal{T}(i)} \alpha_{ij} \cdot \mathbf{h}_j
 \end{equation}
where $\alpha_{ij}$ are the softmax-normalized attention weights and $\tilde{\mathbf{h}}_i$ represents the locally enhanced representation for token $i$. For the first position we set $\tilde{\mathbf{h}}_0=\mathbf{0}$, since $\mathcal{N}(0)=\emptyset$.

The final locally enhanced representation is obtained via residual fusion and linear projection:
\begin{equation}
L_i^{(l)} = h_i^{(l)} + \tilde h_i
\end{equation}
\begin{equation}
\bar L_i^{(l)} = L_i^{(l)} W_L
\end{equation}
where $h_i^{(l)}, \tilde h_i \in \mathbb{R}^{d}$ are the original and locally aggregated representations of token $i$, respectively, and $W_L \in \mathbb{R}^{d \times d'}$ projects them into the query space of the cross-layer memory. The resulting $\bar{\mathbf{L}}^{(l)} \in \mathbb{R}^{n \times d'}$ is the query used in Section~\ref{sec:fusion}. More details are provided in Appendix D.
\subsection{Global Cross-layer Context Fusion}
\label{sec:global}
To overcome the limitation of traditional MoE structures that only select experts based on current layer inputs, we design a cross-layer attention mechanism to fuse global semantic information from historical layers. This branch supplies the keys and values of the routing attention; the query is the local similarity-aware representation of the current layer, constructed in Section~\ref{sec:local}, and the attention itself is carried out in Section~\ref{sec:fusion}.

Specifically, let the input representation of the current $l^{th}$ layer be $\mathbf{H}^{(l)} \in \mathbb{R}^{n \times d}$, where $n$ is the sequence length and $d$ is the hidden dimension. We maintain a historical output cache $\mathcal{M}^{(l)} = \left\{ \mathbf{H}^{(l-1)}, \mathbf{H}^{(l-2)}, \dots, \mathbf{H}^{(l-k)} \right\}$, where $k$ is the number of historical layers to consider, and introduce layer-level index embeddings $\mathbf{E}^{(l-j)}$ for each historical layer. The historical information concatenation form is:
 \begin{equation}
\mathbf{C}^{(l)}=[\mathbf{H}^{(l-1)}+\mathbf{E}^{(l-1)};\ldots;\mathbf{H}^{(l-k)}+\mathbf{E}^{(l-k)}]
 \end{equation}
\begin{equation}
\quad\mathbf{K}^{(l)}=\mathbf{C}^{(l)}\mathbf{W}^K;\quad\mathbf{V}^{(l)}=\mathbf{C}^{(l)}\mathbf{W}^V
 \end{equation}
where $\mathbf{W}^K, \mathbf{W}^V \in \mathbb{R}^{d \times d'}$ are linear transformation matrices for Key and Value, so that $\mathbf{C}^{(l)} \in \mathbb{R}^{kn \times d}$ and $\mathbf{K}^{(l)}, \mathbf{V}^{(l)} \in \mathbb{R}^{kn \times d'}$.

To maintain autoregressive consistency in the cross-layer attention mechanism, we introduce a causal mask $\mathbf{M}_{\text{causal}} \in \mathbb{R}^{n \times kn}$ that constrains the attention patterns. Since $\mathbf{C}^{(l)}$ stacks $k$ historical layers along the sequence axis, a key index $j \in \{0,\dots,kn-1\}$ is \emph{not} itself a sequence position: the $j$-th key is the representation of sequence position
 \begin{equation}
 \pi(j) = j \bmod n
 \end{equation}
taken from historical layer $l-\lfloor j/n \rfloor-1$. The mask is therefore defined on the within-layer position $\pi(j)$ rather than on the absolute key index:
 \begin{equation}
 \mathbf{M}_{\text{causal}}[i,j] = \begin{cases} 
 0 & \text{if } \pi(j) \leq i \\
 -\infty & \text{if } \pi(j) > i
 \end{cases}
 \end{equation}
where $i \in \{0,\dots,n-1\}$ is the query position. Defining the mask this way applies the same constraint to every historical layer: token $i$ may attend to the cached representations of positions $0,\dots,i$ in each of the $k$ layers, and to none of the positions after $i$. Thresholding on the absolute index $j$ instead would leave $\mathbf{H}^{(l-1)}$ partially visible while masking out almost all of $\mathbf{H}^{(l-2)},\dots,\mathbf{H}^{(l-k)}$, which is a window restriction rather than a causal constraint.

Together, $\mathbf{K}^{(l)}$, $\mathbf{V}^{(l)}$ and $\mathbf{M}_{\text{causal}}$ constitute the cross-layer memory of layer $l$: a position-aligned, causally masked summary of how each token has been represented over the preceding $k$ layers.

\begin{table*}[t]
\centering
\small
\newcolumntype{C}[1]{>{\centering\arraybackslash}p{#1}}
\begin{tabular}{l C{1.8cm} C{1.1cm} C{1.1cm} C{1.1cm} C{1.1cm} C{1.1cm} C{1.1cm}}
\toprule
\multicolumn{8}{c}{\textbf{Pre-training Results}} \\
\midrule
Algorithm & Backbone
  & \multicolumn{2}{c}{Enwiki8 (bpc$\downarrow$)}
  & \multicolumn{2}{c}{WikiText-103 (ppl$\downarrow$)}
  & \multicolumn{2}{c}{Params (B)} \\
\midrule
Dense       & \multicolumn{1}{c}{}
  & \multicolumn{2}{c}{1.166}
  & \multicolumn{2}{c}{27.361}
  & \multicolumn{2}{c}{0.054} \\
SMOE        & \multicolumn{1}{c}{}
  & \multicolumn{2}{c}{1.307}
  & \multicolumn{2}{c}{46.956}
  & \multicolumn{2}{c}{0.032} \\
RMOE        & \multicolumn{1}{c}{}
  & \multicolumn{2}{c}{1.293}
  & \multicolumn{2}{c}{39.491}
  & \multicolumn{2}{c}{0.042} \\
SMOE-drop   & \multicolumn{1}{c}{\textbf{Transformer-XL}}
  & \multicolumn{2}{c}{1.262}
  & \multicolumn{2}{c}{39.557}
  & \multicolumn{2}{c}{0.032} \\
HyperRouter & \multicolumn{1}{c}{}
  & \multicolumn{2}{c}{\underline{1.141}}
  & \multicolumn{2}{c}{\underline{26.405}}
  & \multicolumn{2}{c}{0.038} \\
MCF-MOE     & \multicolumn{1}{c}{}
  & \multicolumn{2}{c}{\textbf{1.126}}
  & \multicolumn{2}{c}{\textbf{25.975}}
  & \multicolumn{2}{c}{0.035} \\

\midrule
Algorithm & Backbone
  & \multicolumn{4}{c}{C4 (bpc$\downarrow$)}
  & \multicolumn{2}{c}{Params (B)} \\
\midrule
SMOE        & \multicolumn{1}{c}{}
  & \multicolumn{4}{c}{5.294}
  & \multicolumn{2}{c}{16.376} \\
RMOE        & \multicolumn{1}{c}{}
  & \multicolumn{4}{c}{4.283}
  & \multicolumn{2}{c}{16.444} \\
SMOE-drop   & \multicolumn{1}{c}{\textbf{DeepSeek-MoE}}
  & \multicolumn{4}{c}{\underline{3.058}}
  & \multicolumn{2}{c}{16.376} \\
HyperRouter & \multicolumn{1}{c}{}
  & \multicolumn{4}{c}{4.280}
  & \multicolumn{2}{c}{17.300} \\
MCF-MOE     & \multicolumn{1}{c}{}
  & \multicolumn{4}{c}{\textbf{1.099}}
  & \multicolumn{2}{c}{16.489} \\

\midrule
\multicolumn{8}{c}{\textbf{Fine-tuning Results }} \\
\midrule
Algorithm   & SST2  & QQP   & QNLI  & RTE   & CoLA  & WNLI  & AVG   \\
\midrule
Dense      & 73.61 & 72.25 & 56.63 & 49.26 & 67.02 & {23.44} & {57.04}  \\
SMOE        & 77.08 & 72.98 & 52.79 & 47.43 & 62.40 & 32.81 & 57.58 \\
RMOE        & \underline{79.86} & \underline{74.14} & 56.82 & \underline{52.21} & 61.92 & 20.31 & 57.54 \\
SMOE-drop   & 76.74 & 73.18 & \underline{57.83} & 51.10 & {63.46} & 42.19 & \underline{60.75} \\
HyperRouter & 67.48 & 69.13 & 53.95 & \underline{52.21} & 61.73 & \underline{51.56} & 59.34 \\
MCF-MOE     & \textbf{80.21} & \textbf{74.55} & \textbf{61.61} & \textbf{54.41} & \textbf{69.13} & \textbf{56.25} & \textbf{66.03} \\
\bottomrule
\end{tabular}
\caption{Pre-training and fine-tuning results of various MoE models.
  Best results are highlighted in \textbf{bold},
  second-best results are \underline{underlined}. For convenience, we use “SMOE-drop” to denote SMOE-Dropout. }
\label{tab:moe_results}
\end{table*}

\subsection{Multi-level Context Fusion for Gating}
\label{sec:fusion}
The two levels of context meet in a single cross-attention: the local similarity-aware representation $\bar{\mathbf{L}}^{(l)}$ of Section~\ref{sec:local} acts as the query, and the cross-layer memory of Section~\ref{sec:global} supplies the keys and values. Writing $\mathbf{Q}^{(l)} = \bar{\mathbf{L}}^{(l)}$, the attention score of query position $i$ against key $j$ is
 \begin{equation}
 \text{score}_{i,j} = \frac{\mathbf{Q}^{(l)}_i \cdot (\mathbf{K}^{(l)}_j)^\top}{\sqrt{d^{\prime}}} + \mathbf{M}_{\text{causal}}[i,j]
 \end{equation}
where keys with $\pi(j) > i$ are masked to prevent information leakage from future tokens, resulting in zero attention weights after softmax normalization. This fusion lets local context determine how each token retrieves from its cross-layer history, allowing routing to adapt to local patterns. The fused routing representation is therefore:
\begin{equation}
\mathbf{z}^{(l)} = \mathrm{softmax}\left(\frac{\mathbf{Q}^{(l)}(\mathbf{K}^{(l)})^\top}{\sqrt{d^{\prime}}}+\mathbf{M}_{\mathrm{causal}}\right)\mathbf{V}^{(l)} 
\end{equation}

This fused representation serves as the input to the expert projection layer to generate routing logits. The expert selection scores are computed as:
 \begin{equation}
 \mathbf{g}^{(l)}=\mathbf{z}^{(l)}\mathbf{W}_{expert}
 \end{equation}
where $\mathbf{g}^{(l)}\in\mathbb{R}^{N}$ are the expert selection logits and $\mathbf{W}_{expert} \in \mathbb{R}^{d' \times N}$ is the expert projection matrix with $N$ being the total number of experts. The final MoE output selects the top-$K$ experts based on these scores:
 \begin{equation}
 \mathrm{M}\mathrm{O}\mathrm{E}(\mathbf{h}_i^{(l)})=\sum_{j\in\mathcal{T}_K(\mathbf{g}_i)}\mathrm{s}\mathrm{o}\mathrm{f}\mathrm{t}\mathrm{m}ax(\mathbf{g}_i)_j\cdot E_j(\mathbf{h}_i^{(l)})
 \end{equation}
where $\mathcal{T}_K(\cdot)$ represents the top-$K$ function, and $E_j(\cdot)$ represents the $j$-th expert. 

MCF-MOE integrates multi-level contextual signals to capture fine-grained token dependencies and offers two benefits: a global context fusion module that aggregates hierarchical cues from previous layers to stabilize expert selection across similar tokens, and a local similarity–aware routing that concentrates on semantically relevant tokens to balance expert usage and mitigate representation collapse.

\section{Experiments}

\subsection{Experiment Settings}
\subsubsection{Dataset}
We adopt a two-stage pretrain–finetune paradigm. In the pre-training stage, we use Enwiki8 \cite{mahoney2011large} and WikiText-103 \cite{merity2016pointer} to evaluate character-level and word-level language modeling under a controlled Transformer-XL setting. In addition, to assess scalability in large-scale pre-training, we conduct experiments on the C4 dataset using decoder-only backbones. In the fine-tuning stage, we evaluate general language understanding on six representative datasets: SST-2, QQP, QNLI, RTE, CoLA, and WNLI. Further dataset details are provided in Appendix~A.

\subsubsection{Implementation Details}
Most experiments are conducted on four NVIDIA A800 GPUs. We first adopt Transformer-XL as the backbone to isolate the effects of MoE routing within a standard Transformer architecture on Enwiki8 and Wikitext-103. To further evaluate generalization and scalability, we transfer our method to DeepSeek-MoE, a larger decoder-only backbones, and conduct experiments on the larger C4 dataset. All inputs are tokenized with the corresponding tokenizer and truncated or padded to a maximum sequence length of 512 tokens. Apart from tokenizer-specific decoding, the preprocessing pipeline (filtering, sampling, and sequence length) is identical across backbones. The details of preprocessing are provided in Appendix B.

Following standard practice, we report bits-per-character (bpc) on Enwiki8 for character-level language modeling, and perplexity (ppl) on Wikitext-103 for word-level language modeling. For large-scale pre-training on C4, we use bpc as the evaluation metric for decoder-only models. 
All experiments employ a top-2-of-16 routing scheme, with four out of eight layers implemented as MoE layers. Full training procedures and hyperparameter settings are provided in Appendix~B.

\subsection{Main Results}

Table~\ref{tab:moe_results} summarizes the performance of dense and MoE-based models in both pre-training and downstream fine-tuning settings.

\textbf{Pre-training Performance.}
On Enwiki8 and WikiText-103 with the Transformer-XL backbone, MCF-MOE achieves the best results among all methods. Despite using only 34.8M parameters, it outperforms the dense Transformer-XL model (53.5M) and significantly improves over strong MoE baselines, reducing bpc by over 0.13 and perplexity by more than 10 compared to SMOE-Dropout, indicating more effective routing under limited capacity.

To evaluate scalability, we further conduct experiments on a decoder-only LLM with the DeepSeek-MoE backbone trained on C4. At convergence, MCF-MOE consistently achieves the lowest bpc across both architectures.
Under the larger DeepSeek-MoE backbone ($\approx$16B parameters), it further improves bpc to \textbf{1.099}, compared to 3.058 and 4.283 for SMOE-Dropout and RMOE, respectively. These gains are achieved with comparable parameter counts, suggesting that improvements mainly stem from more effective routing.

\textbf{Fine-tuning Performance.}
We further evaluate downstream transferability by fine-tuning each model on six downstream language modeling tasks (SST-2, QQP, QNLI, RTE, CoLA, and WNLI). As shown in Table~\ref{tab:moe_results}, MCF-MOE achieves the highest average accuracy (66.03\%), outperforming all baselines, with improvements of 6.69\% over HyperRouter and 9.79\% over RMOE. It also surpasses the dense model across almost all tasks, indicating that the gains primarily arise from improved routing.

Task-wise, MCF-MOE achieves the best performance across all six downstream tasks, demonstrating consistent gains across sentiment analysis, paraphrase detection, linguistic acceptability, and natural language inference. The improvements are particularly pronounced on QNLI, CoLA, and WNLI, while MCF-MOE also maintains clear advantages on SST-2, QQP, and RTE. These results demonstrate that multi-level context fusion improves generalization across diverse language understanding tasks rather than benefiting only specific task types.

\subsection{Ablation Studies}
\subsubsection{Ablation experiments of different modules}
To assess the contribution of each component in MCF-MOE, we perform ablation studies on both Enwiki8 and Wikitext-103 for 20k training steps (Table \ref{tab:Ablation}). We progressively remove modules from the full model and observe consistent degradation across the two datasets.



\begin{table}[h]
\small
\centering
\resizebox{1\columnwidth}{!}{%
\begin{tabular}{lcccc}
\toprule
\textbf{Model Variant} & \multicolumn{2}{c}{\textbf{Enwiki8}} & \multicolumn{2}{c}{\textbf{Wikitext-103}} \\
 & \textbf{bpc} & \boldmath$\Delta$ & \textbf{ppl} & \boldmath$\Delta$ \\
\midrule
MCF-MOE (Full) & 1.184 & -- & 27.837 & -- \\
\midrule
\textbf{w/o Global Context} & 1.219 & +0.035 & 28.121 & +0.284 \\
\quad \textit{w/o Cross attention} & 1.209 & +0.025 & 28.045 & +0.208 \\
\quad \textit{w/o Causal Mask} & 1.210 & +0.026 & 28.098 & +0.261 \\
\midrule
\textbf{w/o Local Context} & 1.205 & +0.032 & 28.533 & +0.696 \\
\midrule
\textbf{Standard MOE} & 1.347 & +0.163 & 29.085 & +1.248 \\
\bottomrule
\end{tabular}%
}
\caption{Ablation study of MCF-MOE on Enwiki8 and Wikitext-103 using bits-per-character (bpc) and perplexity as the evaluation metric. Lower is better.}
\label{tab:Ablation}
\end{table}


Removing the global context fusion module leads to a clear performance drop on both datasets, increasing the bpc from 1.184 to 1.219 on Enwiki8 and the perplexity from 27.837 to 28.121 on Wikitext-103. A finer-grained ablation further shows that eliminating cross-attention causes the larger degradation, while disabling the causal mask yields slightly smaller but still consistent declines, indicating that cross-attention contributes more to global semantic aggregation, with causal masking providing complementary gains.
Ablating the local context perception module also consistently hurts performance, suggesting that fine-grained token-level cues are important for effective expert specialization.
Finally, removing both global and local components—reducing the model to a standard MoE gating strategy—produces the largest degradation, confirming that global and local cues are complementary: global fusion stabilizes routing through multi-layer semantics, while local fusion improves discrimination via token-level patterns.

\subsubsection{Ablation experiments of sensitive parameters}
We conduct ablation experiments to analyze the impact of fusion strategy, model depth, and historical context window $k$ on Enwiki8. Results are shown in Table~\ref{tab:ablation_all}.
\begin{table}[h]
\centering
\resizebox{1\columnwidth}{!}{%
\begin{tabular}{llcc}
\toprule
\textbf{Category} & \textbf{Setting} & \textbf{Val} & \textbf{Test} \\
\midrule
\multirow{5}{*}{Fusion Strategy}
& Concat              & 1.233          & 1.199          \\
& Element-wise Add    & 1.235          & 1.203          \\
& Simple Gating       & 1.237          & 1.203          \\
& Multi-Gating        & 1.236          & 1.204          \\
& Cross-Attn (Ours)   & \textbf{1.227} & \textbf{1.184} \\
\midrule
\multirow{4}{*}{Model Depth}
& 4-layers            & 1.302          & 1.271          \\
& 8-layers (Ours)     & \textbf{1.227} & \textbf{1.184} \\
& 12-layers           & 1.272          & 1.233          \\
& 16-layers           & 1.272          & 1.233          \\
\midrule
\multirow{5}{*}{Window $k$}
& 0                   & 1.242          & 1.219          \\
& 1                   & 1.239          & 1.208          \\
& 3 (Ours)            & \textbf{1.227} & \textbf{1.184} \\
& 5                   & 1.229          & 1.194          \\
& 8                   & 1.294          & 1.255          \\
\bottomrule
\end{tabular}%
}
\caption{Routing design analysis on fusion strategy, model depth, and history window $k$ on Enwiki8.}
\label{tab:ablation_all}
\end{table}

\textbf{Fusion Strategy.}
We compare cross-attention with concatenation, element-wise addition, and gating variants. Cross-attention performs best, suggesting that routing benefits from structured interactions between global and local contexts rather than simple feature mixing.

\textbf{Model Depth.}
Among models with 4, 8, 12, and 16 layers, the 8-layer model performs best. Increasing depth further provides no additional benefit, indicating that model depth cannot compensate for suboptimal routing.

\textbf{History Window ($k$).}
Performance improves as $k$ increases from 0 to 3 but declines with larger windows, as excessive history may introduce noise. We therefore use $k=3$. Further details are provided in Appendix~E.


\subsection{Efficiency Analysis}

\paragraph{Inference Latency}
Table~\ref{tab:inference_efficiency} reports end-to-end inference latency across different sequence lengths. At 512 tokens, MCF-MOE is slightly slower than Dense and SMOE-Dropout, reflecting the additional cost of multi-level context fusion. However, MCF-MOE scales more favorably as the sequence length increases and achieves the lowest latency from 1,024 to 4,096 tokens. At 4,096 tokens, it remains the fastest method, with a particularly clear advantage over RMOE and HyperRouter. These results demonstrate the practical efficiency of MCF-MOE for long-context inference.

\paragraph{Inference Memory}

Table~\ref{tab:inference_efficiency} also reports peak inference memory across sequence lengths. All methods exhibit similar memory scaling, increasing from approximately 0.8\,GB at 512 tokens to 3.1--3.2\,GB at 4,096 tokens. MCF-MOE achieves the lowest memory usage at 512 tokens and remains within 8.8\% of the most memory-efficient method at all sequence lengths. At 4,096 tokens, it consumes 3.17\,GB, compared with 3.11\,GB for HyperRouter and 3.20\,GB for RMOE. These results indicate that the global and local context fusion modules introduce only modest memory overhead while maintaining a memory profile comparable to existing routing methods.

\begin{table*}[t]
\centering
\small
\setlength{\tabcolsep}{3pt}
\resizebox{\textwidth}{!}{%
\begin{tabular}{lrrrrrrrr}
\toprule
\multirow{2}{*}{\textbf{Model}} &
\multicolumn{4}{c}{\textbf{Inference Latency (ms) $\downarrow$}} &
\multicolumn{4}{c}{\textbf{Peak Memory (MB) $\downarrow$}} \\
\cmidrule(lr){2-5}\cmidrule(lr){6-9}
& \textbf{512} & \textbf{1,024} & \textbf{2,048} & \textbf{4,096}
& \textbf{512} & \textbf{1,024} & \textbf{2,048} & \textbf{4,096} \\
\midrule
Dense       & \textbf{373.51} & 2373.18 & 6476.54 & 18670.08
            & 819.48 & \underline{960.98} & \underline{1498.20} & 3129.36 \\
SMOE        & 1052.43 & 2272.66 & 5170.95 & \underline{16577.62}
            & 814.12 & 1002.13 & 1515.77 & \underline{3125.23} \\
RMOE        & 2491.48 & 4983.92 & 8582.39 & 22170.06
            & 868.74 & 1057.36 & 1591.29 & 3196.50 \\
SMOE-Drop   & \underline{634.05} & \underline{1569.54} & 5114.14 & 17042.39
            & \underline{811.03} & 1019.78 & 1528.06 & 3156.15 \\
HyperRouter & 1133.29 & 2273.26 & \underline{5072.63} & 18744.83
            & 842.83 & \textbf{946.93} & \textbf{1484.64} & \textbf{3108.16} \\
MCF-MOE     & 696.19 & \textbf{1080.32} & \textbf{4678.29} & \textbf{16002.83}
            & \textbf{807.62} & 1030.12 & 1516.60 & 3174.66 \\
\bottomrule
\end{tabular}
}
\caption{End-to-end inference latency and peak inference memory across sequence lengths. Lower is better. Best results are in \textbf{bold}, and second-best results are \underline{underlined}.}
\label{tab:inference_efficiency}
\end{table*}



\subsection{Observations}
\subsubsection{MCF-MOE Promotes Semantically Coherent Expert Specialization}
We measure whether semantically related tokens are routed to the same experts using normalized mutual information (NMI). Token occurrences are grouped into $K$ semantic clusters using representations from a frozen external encoder. Let $G$ denote the cluster label and $E$ the top-1 expert assignment. We compute
\begin{equation}
    \operatorname{NMI}(G,E)
    =
    \frac{2I(G;E)}{H(G)+H(E)}.
\end{equation}
Higher values indicate that expert assignments preserve more semantic-group information beyond that explained by marginal expert utilization.
To account for non-zero NMI caused by finite samples and imbalanced expert usage, we report the lift over a permutation baseline:
\begin{equation}
    \Delta\operatorname{NMI}
    =
    \operatorname{NMI}(G,E)
    -
    \mathbb{E}_{\pi}
    \left[\operatorname{NMI}(G,\pi(E))\right],
\end{equation}
where $\pi(E)$ randomly permutes expert assignments within each MoE layer while preserving their marginal distribution.
\begin{figure}[h]
    \centering
    \includegraphics[width=\linewidth]{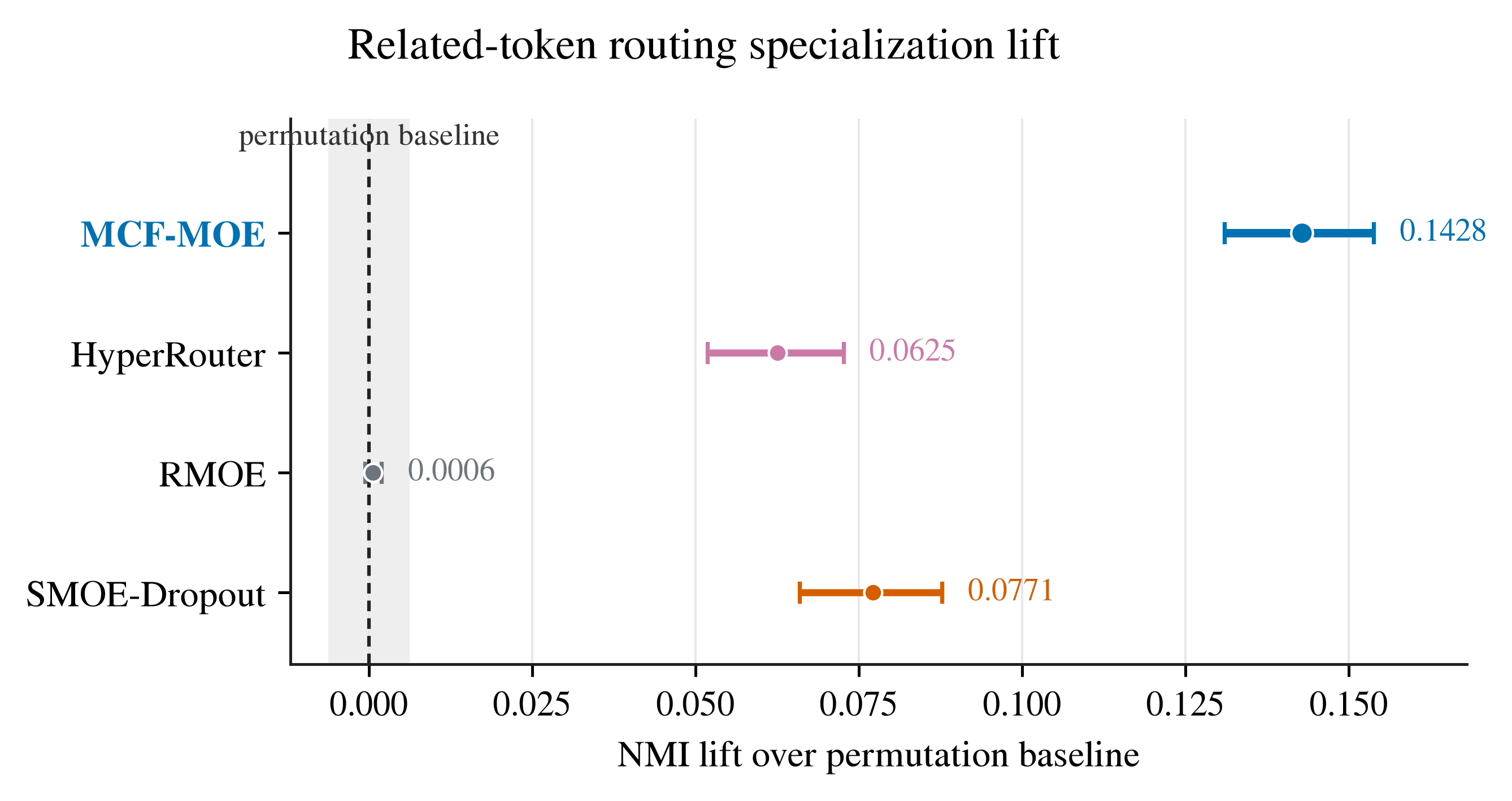}
    \caption{
        Related-token routing specialization measured by NMI lift over a permutation baseline.
        Error bars represent 95\% confidence intervals across random seeds.
    }
    \label{fig:nmi_lift}
\end{figure}

As shown in Figure~\ref{fig:nmi_lift}, MCF-MOE obtains the largest NMI lift ($0.1428$), compared with SMOE-Dropout ($0.0771$), HyperRouter ($0.0625$), and RMOE ($0.0006$). Thus, its routing decisions retain substantially more information about token semantics than expected from expert usage alone. Together with the performance degradation observed after removing either context module (Table~\ref{tab:Ablation}), this result suggests that multi-level context encourages semantically coherent expert specialization and contributes to the downstream gains.

\subsubsection{MCF-MOE achieves more balanced and stable expert utilization across layers}

\begin{figure}[h]
\centering
\includegraphics[width=0.45\textwidth]{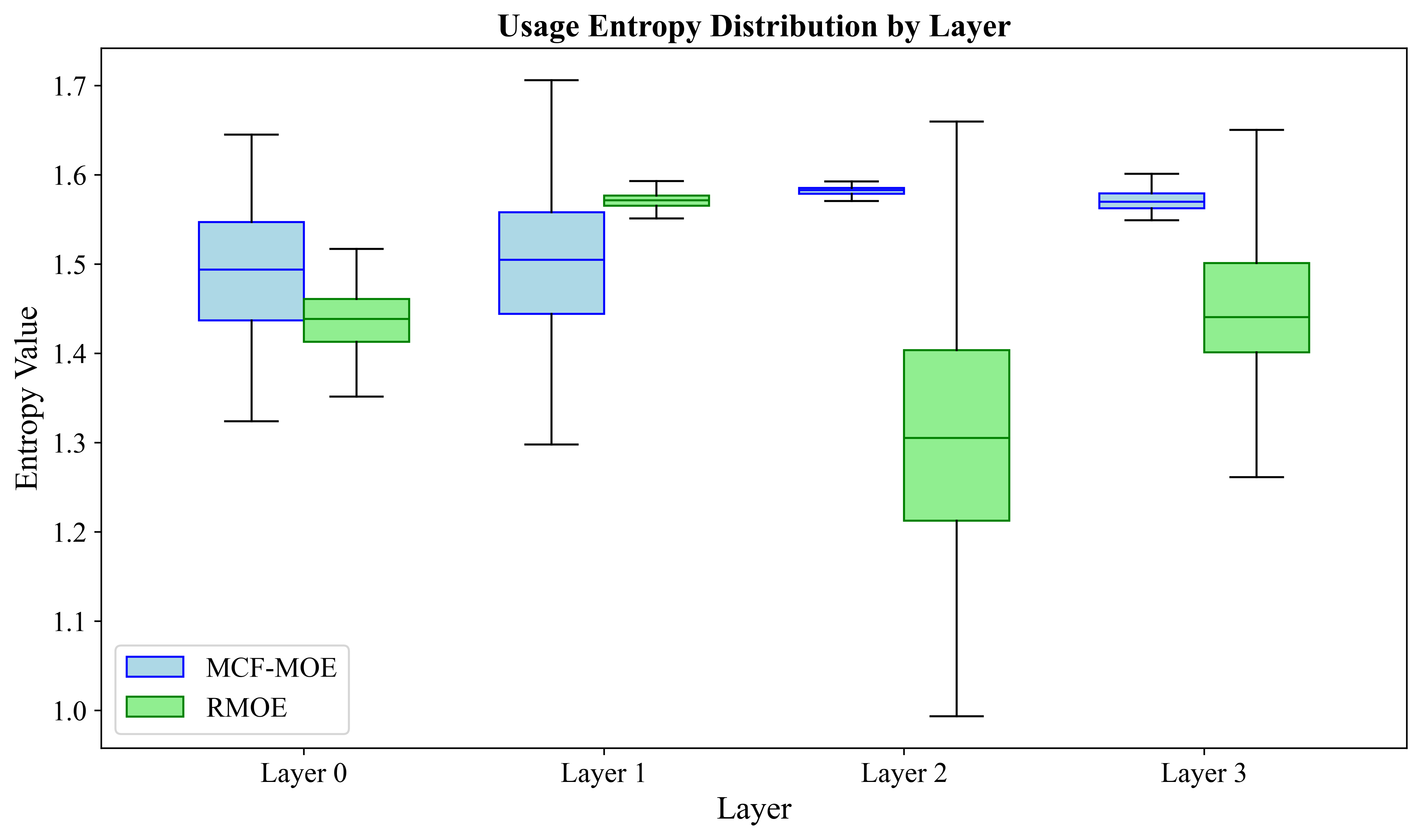} 
\caption{Layer-wise comparison of expert usage entropy between MCF-MOE and RMOE. 
}                                            
\label{Fig:Entropy usage}
\end{figure}

To compare routing balance across layers, we report the expert usage entropy of MCF-MOE and RMOE, where higher entropy indicates more uniform expert utilization and improved capacity usage. We choose RMOE as the baseline since it incorporates cross-layer information via a GRU-based router. Figure~\ref{Fig:Entropy usage} shows the entropy results for the four MoE layers.

At Layer 0, RMOE exhibits slightly higher entropy due to its router operating on raw inputs, while MCF-MOE shows larger variance caused by noisy similarity signals at shallow layers. From Layer 1 onward, MCF-MOE consistently achieves higher and more stable entropy, whereas RMOE exhibits increased variance and reduced entropy. In deeper layers (Layers 2–3), MCF-MOE maintains balanced expert utilization, while RMOE shows stronger routing concentration.
These results demonstrate the robustness of MCF-MOE in sustaining balanced expert usage across model depth.

\subsubsection{MCF-MOE Exhibits Structured Expert Collaboration and  Task Specialization
}

\begin{figure}[h]
\centering
\includegraphics[width=0.5\textwidth]{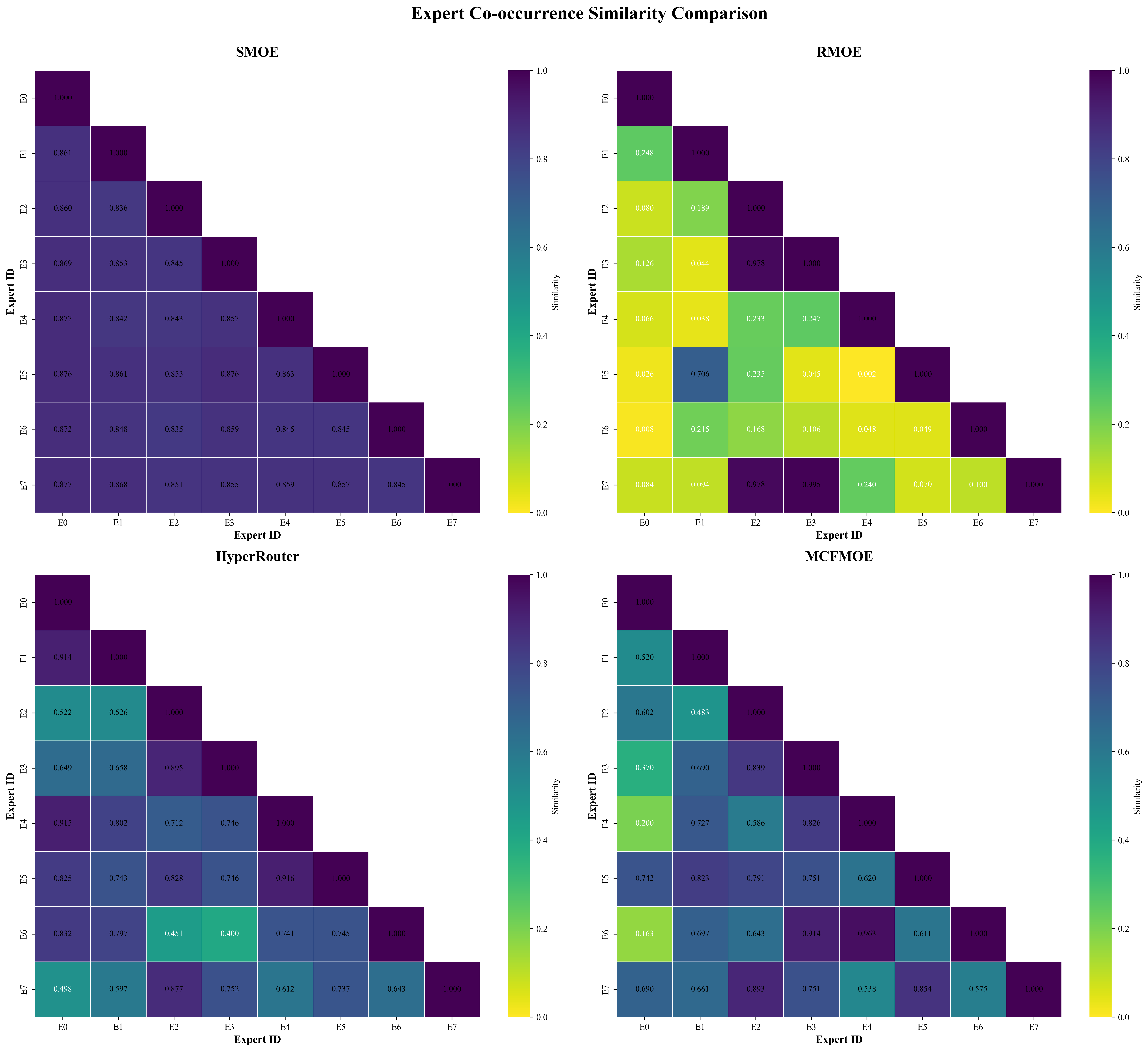} 
\caption{Expert Occurrence Similarity Matrix of MCF-MOE, which illustrates the co-activation probability between expert pairs in MCF-MOE when processing the same token. }
\label{heatmap}
\end{figure}

To analyze how routers shape expert interaction, we compute expert co-occurrence similarity, measuring how often two experts are jointly activated for the same token. This indicates whether a router promotes meaningful collaboration or collapses into redundant co-activation, and whether it preserves specialization.

As shown in Figure \ref{heatmap}, MCF-MOE shows a non-uniform co-occurrence distribution: pairs like (3,6), (3,4), and (2,3) co-activate frequently, suggesting collaborative processing of semantically related tokens, likely due to multi-granular contextual fusion. In contrast, pairs such as (0,4) and (0,6) co-activate less, indicating independent specialization and reduced redundancy.

HyperRouter exhibits consistently high co-occurrence (>0.70), reflecting limited specialization, while SMOE shows nearly uniform co-activation (>0.84) across pairs. RMOE produces sparse and irregular activations, with few co-firing pairs and many disconnected, indicating fragmented collaboration.

Overall, only MCF-MOE achieves both targeted collaboration within semantically coherent expert groups and clear specialization across groups.


\section{Conclusion}
In this work, we revisit expert routing in Mixture-of-Experts models from the perspective of routing context specification. We show that relying on single-layer representations or compact memory states can limit semantically coherent and stable expert selection across layers. To address this issue, we propose MCF-MOE, which reformulates expert routing as a decision over multi-level contextual evidence by jointly modeling global semantic alignment and local token-level stability. Extensive experiments  demonstrate consistent improvements over strong MoE baselines and robustness to different parameter settings. We hope this perspective highlights the importance of explicit routing context modeling and offers a principled direction for future MoE routing design.

\section*{Limitations}
We further show that MCF-MOE achieves strong and consistent performance on large-scale decoder-only language models, demonstrating its scalability and architectural compatibility beyond conventional Transformer settings. Despite its effectiveness, MCF-MOE has several limitations. First, it assumes that representations across layers can be effectively aligned for routing, which may become less reliable in very deep decoder-only models due to increasing semantic divergence. Second, MCF-MOE introduces additional hyperparameters for multi-level context fusion, making routing behavior sensitive to these choices and requiring careful tuning across different model scales and tasks. We leave these directions for future work.


\bibliography{main}
The appendix of our work is organized as follows.
\appendix 
\section{Dataset}
\subsection{Pre-training Dataset}
We employ three widely-used datasets for pre-training: Enwiki8, WikiText-103 and C4.

Enwiki8 is a character-level dataset containing the first 100MB of English Wikipedia text. It is commonly used for evaluating compression and character prediction performance, with bits per character (bpc) as the metric—lower bpc indicates better compression. Its compact and highly compressible nature makes it ideal for testing lightweight models under limited-context settings.

WikiText-103 is a large-scale word-level corpus derived from high-quality Wikipedia articles, preserving punctuation and document structure. It is suitable for modeling long-range dependencies, with perplexity (ppl) as the evaluation metric—lower ppl implies stronger language modeling ability.

C4 (Colossal Clean Crawled Corpus) is a large-scale, web-crawled text corpus derived from Common Crawl with extensive filtering and deduplication. It spans a wide variety of topics and writing styles, providing diverse and realistic web text suitable for large-scale language model pre-training. Its size and diversity make it ideal for evaluating model scalability, generalization, and stability in large-context and high-capacity settings. We construct the training corpus by sampling from the \textit{RealNews} split of the C4 dataset, which is a news-domain filtered subset widely used in language model pretraining. The RealNews training split contains approximately 13.8 million documents (about 36.9 GiB of raw text). From this split, we uniformly sample a fixed subset that is shared across all routing methods to ensure consistent training data across experiments.
\subsection{Finetune dataset}
For the fine-tuning stage, we adopt six representative  datasets—SST-2, QQP, QNLI, and RTE, CoLA, and WNLI—for downstream evaluation, which span sentiment classification, paraphrase detection, and natural language inference tasks. 
\begin{itemize}
    \item \textbf{SST‑2}: The Stanford Sentiment Treebank II (SST‑2) is a binary sentiment classification task on single sentences from movie reviews. With approximately 67K training examples, models predict whether each sentence is positive or negative. Performance is assessed via accuracy, reflecting sentiment prediction correctness.
\end{itemize}
\begin{itemize}
    \item \textbf{QQP}: Quora Question Pairs (QQP) is a paraphrase detection task that determines if two Quora questions are semantically equivalent. It includes roughly 364K training pairs. Evaluation uses accuracy and F1 score to capture correctness and class balance.
\end{itemize}
\begin{itemize}
    \item \textbf{QNLI}: Question‑answering Natural Language Inference (QNLI) converts SQuAD‑style question–sentence pairs into a binary entailment task: does the sentence contain the answer? It provides around 105K training pairs, and uses accuracy as its evaluation metric.
\end{itemize}
\begin{itemize}
    \item \textbf{RTE}: The Recognizing Textual Entailment (RTE) task is a binary natural language inference dataset consolidated from multiple RTE challenges. It is relatively small, with about 2.5K training examples, and evaluated via accuracy to judge whether one sentence entails the other.
\end{itemize}
\begin{itemize}
    \item \textbf{CoLA}: The Corpus of Linguistic Acceptability (CoLA) evaluates a model’s ability to judge the grammatical acceptability of English sentences. Each example is labeled as either grammatically acceptable or unacceptable. The dataset contains about 8.5K training sentences and is evaluated using the Matthews correlation coefficient (MCC), which is designed for binary classification with class imbalance.
\end{itemize}

\begin{itemize}
    \item \textbf{WNLI}: The Winograd Natural Language Inference (WNLI) task is derived from the Winograd Schema Challenge and focuses on pronoun resolution and commonsense reasoning. Given a premise–hypothesis pair involving ambiguous coreference, the model must determine whether the hypothesis is entailed by the premise. WNLI is a low-resource dataset with only a few hundred training examples and is evaluated using accuracy. 
\end{itemize}

\subsection{Data Preprocessing}

\paragraph{Tokenization and Sequence Processing.}
To avoid introducing confounding factors caused by mismatched vocabularies, each model adopts the tokenizer corresponding to its backbone architecture. Specifically, Transformer-XL uses its original subword tokenizer, and DeepSeek-MoE uses the tokenizer released with the DeepSeek base model. All input texts are tokenized using the corresponding tokenizer and truncated or padded to a maximum sequence length of 512 tokens. Apart from tokenizer-specific tokenization, the remaining preprocessing steps (data filtering, sampling strategy, and sequence length configuration) are identical across all backbones.

The preprocessing configurations for each backbone are summarized in Table~\ref{tab:c4_preprocess}.

\begin{table}[h]
\centering
\small
\resizebox{1\columnwidth}{!}{%
\begin{tabular}{llll}
\toprule
\textbf{Backbone} & \textbf{Tokenizer} & \textbf{Type} & \textbf{Max Len} \\
\midrule
Transformer-XL & Transformer-XL tokenizer & Word-level & 512 \\
DeepSeek-MoE   & DeepSeek tokenizer       & BPE        & 512 \\
\bottomrule
\end{tabular}%
}
\caption{Preprocessing configuration for C4-based experiments.}
\label{tab:c4_preprocess}
\end{table}

\section{Experiment Configuration}
\subsection{Pre-training configuration}
During the pre-training stage, we perform dataset-specific pretraining based on the Transformer-XL architecture. The training is conducted four NVIDIA A800 GPU. Table~\ref{tab:config} summarizes the detailed configurations used in the main pretraining experiments.
\begin{table}[h]
\centering
\small 
\setlength{\tabcolsep}{2pt} 
\renewcommand{\arraystretch}{0.9} 
\begin{tabular}{lccccc}
\toprule
Layer & Optimizer & Lr & Batch Size & Steps & Experts \\
\midrule
8 & Adam & 0.0007 & 48 & 80k & 16 \\
\bottomrule
\end{tabular}
\caption{Experimental configurations for pre-training.}
\label{tab:config}
\end{table}

\begin{table}[h]
\centering
\small 
\setlength{\tabcolsep}{2pt} 
\renewcommand{\arraystretch}{0.9} 
\begin{tabular}{lccccc}
\toprule
Layer & Optimizer & Lr & Batch Size & Steps & Experts \\
\midrule
8 & Adam & 0.0001 & 48 & 4k & 16 \\
\bottomrule
\end{tabular}
\caption{Experimental configurations for finetuning.}
\label{tab:finetune_config}
\end{table}
\subsection{Finetune configuration}

During the fine-tuning stage, we further adapt the pretrained model to downstream tasks using six GLUE datasets: SST-2, QQP, QNLI, and RTE. All fine-tuning experiments are conducted using four A800 GPU. Table \ref{tab:finetune_config} summarizes the detailed configurations used in the fine-tuning stage.

\section{Baseline models}

We select the most relevant state-of-the-art MoE baseline models for comparison in our main experiments. Below is a brief introduction to each baseline:

\begin{itemize}
  \item \textbf{SMOE}: The standard Sparse Mixture-of-Experts model leverages learned softmax-based routing to dynamically assign tokens to a subset of experts, supporting scalable computation and conditional capacity allocation.
  \end{itemize}
  \begin{itemize}
  \item \textbf{SMOE‑Dropout}: A plug-and-play routing scheme that freezes a randomly initialized router and progressively increases the Top-$k$ expert selection during training, yielding a self-slimmable architecture with flexible expert activation and robust performance across various routing budgets.
  \end{itemize}
  \begin{itemize}
  \item \textbf{HyperRouter}: Combines a fixed hypernetwork and trainable layer embeddings to generate layer-specific routing parameters, achieving a dynamic and semantically adaptive routing strategy with improved stability across layers.
  \end{itemize}
  \begin{itemize}
  \item \textbf{RMOE}: Utilizes a GRU-based memory module to model expert activation trajectories across layers as sequential patterns, enabling enhanced information flow and promoting semantic alignment in routing decisions.
\end{itemize}



\section{Additional Explanation of Local Similarity-Aware Context Fusion for MoE Routing}

In this section, we summarize the key formulation and rationale behind our local similarity-aware context fusion module, which explicitly structures routing context rather than performing standard representation mixing.

\subsection{Self-Attention vs MoE Routing}

Self-attention computes contextualized token representations:
\begin{align}
\mathbf{h}_i^{*} &= \sum_j \alpha_{ij} \mathbf{v}_j, \\
\alpha_{ij} &= \text{softmax}(q_i^\top k_j),
\end{align}
where the goal is representation aggregation.

In contrast, MoE routing performs discrete expert selection:
\begin{equation}
g_i = \text{TopK}(\text{softmax}(W_r \mathbf{h}_i)),
\end{equation}
where the routing decision depends only on the current token embedding $\mathbf{h}_i$. Vanilla attention does not explicitly model structured neighborhood context for routing.

\subsection{Local Similarity-Aware Context Fusion}

Our module constructs an explicit local neighborhood set:
\begin{equation}
\mathcal{N}_i = \text{TopK}_j \, \text{Sim}(\mathbf{h}_i, \mathbf{h}_j), \quad i-r \le j \le i-1,
\end{equation}
and forms a routing-aware context:
\begin{equation}
\tilde{\mathbf{h}}_i = f\Big(\mathbf{h}_i, \frac{1}{|\mathcal{N}_i|} \sum_{j \in \mathcal{N}_i} \mathbf{h}_j \Big),
\end{equation}
where $f$ denotes a fusion function. This explicitly injects local structural coherence into routing decisions.







\section{Algorithm Framework}
To address the limitations of single-layer routing decisions in sparse MOE models, we propose MCF-MOE, a unified routing framework that incorporates Multi-layer Context Fusion to enhance expert selection. 
The detailed algorithmic framework is summarized in Algorithm \ref{alg:mcf-moe}.

\begin{table*}[t]
\centering
\small 
\setlength{\tabcolsep}{7pt} 
\renewcommand{\arraystretch}{1} 
\begin{tabular}{c|cccc|cccc}
\toprule
\multirow{2}{*}{$K$} & \multicolumn{4}{c|}{\textbf{Enwiki8 (bpc)}} & \multicolumn{4}{c}{\textbf{Wikitext-103 (ppl)}} \\
 & HyperRouter & SMOE-Drop & RMOE & MCF-MOE & HyperRouter & SMOE-Drop  & RMOE & MCF-MOE \\
\midrule
1  & 1.144 & 1.848 & 1.652 & \textbf{1.133} & \textbf{27.678} & 79.253 & 51.020 & 28.012 \\
2  & 1.141 & 1.262 & 1.481 & \textbf{1.126} & 26.405 & 39.557 & 41.193 & \textbf{25.975} \\
4  & 1.141 & 1.179 & 1.494 & \textbf{1.125} & 26.541 & 32.704 & 42.533 & \textbf{26.087} \\
8  & 1.140 & 1.155 & 1.530 & \textbf{1.125} & 27.128 & 30.710 & 46.399 & \textbf{26.918} \\
16 & 1.140 & 1.149 & 1.604 & \textbf{1.125} & 28.053 & 30.722 & 54.036 & \textbf{28.012} \\
\bottomrule
\end{tabular}
\caption{Performance comparison of different MoE models on Enwiki8 (measured in bpc) and Wikitext-103 (measured in ppl) under varying top-$K$ settings. Lower values indicate better performance.}
\label{tab:moe_k_comparison}
\end{table*}
\begin{algorithm}[H]
\small 
\setlength{\lineskip}{0pt}
\setlength{\lineskiplimit}{0pt}
\setlength{\abovecaptionskip}{3pt} 
\setlength{\belowcaptionskip}{0pt}
\caption{MCF-MOE: Multi-layer Context Fusion MoE}
\label{alg:mcf-moe}
\textbf{Input}: Current layer input $X_l \in \mathbb{R}^{B \times T \times D}$, layer index $l$\\
\textbf{Parameter}: Number of experts $E$, top-$K$ selection, local radius $r$, local neighbors $k_{\text{loc}}$\\
\textbf{Output}: Expert indices $I$ and routing weights $W$
\begin{algorithmic}[1]
\STATE Initialize historical context $\mathcal{H} = \{H_{l-k}, \dots, H_{l-1}\}$
\IF{$l = 0$ or $|\mathcal{H}| = 0$}
    \STATE $\text{logits} = \text{Linear}(X_l)$
\ELSE
    \STATE \textbf{// Local Similarity Module (query)}
    \STATE $S = X_l \cdot X_l^T$
    \STATE $M_{local}[i,j] = 0$ if $i-r \le j \le i-1$, else $-\infty$
    \FOR{each token $i$}
        \STATE Select top-$k_{\text{loc}}$ neighbors of $S[i,:] + M_{local}[i,:]$
        \STATE $C_{local}[i] = X_l[i] + \sum_j \text{weight}_{ij} \cdot X_l[j]$
    \ENDFOR
    \STATE \textbf{// Cross-layer Context Module (key/value)}
    \STATE $H_{concat} = \text{Concat}(\mathcal{H}) + \text{LayerEmbedding}(l-k{:}\,l-1)$
    \STATE $M_{global}[i,j] = 0$ if $(j \bmod T) \le i$, else $-\infty$
    \STATE \textbf{// Context Fusion and Expert Routing}
    \STATE $X_{fused} = \text{CrossAttention}(C_{local},\, H_{concat},\, M_{global})$
    \STATE $\text{logits} = \text{ExpertProjection}(X_{fused})$
\ENDIF
\STATE $I, W = \text{TopK}(\text{softmax}(\text{logits}/\tau), K)$
\STATE Update $\mathcal{H} \leftarrow \mathcal{H} \cup \{X_l\}$, retaining the most recent $k$ layers
\STATE \textbf{return} $I, W$
\end{algorithmic}
\end{algorithm}
\section{Parameter Analysis}
\subsection{Local Context Sensitivity Analysis}

We analyze the sensitivity of MCF-MOE to the hyperparameters of the local similarity-aware routing module, including the local radius $r$ and the number of selected neighbors (Top-$k$). As shown in Table~\ref{tab:local_param}, MCF-MOE exhibits stable performance across a range of configurations, indicating that the proposed routing strategy is not overly sensitive to specific hyperparameter choices.

When fixing the local radius to $r=16$, selecting a moderate number of neighbors ($\text{Top-}k=8$) yields the best performance. Using too few neighbors limits the amount of contextual information available for routing, while overly large $k$ may introduce noisy or less relevant tokens, slightly degrading performance.

Similarly, increasing the local radius from 16 to 32 does not lead to further improvements, suggesting that most informative local dependencies are already captured within a relatively compact neighborhood. Overall, these results support the design choice of using a moderate local window and Top-$k$ selection to balance contextual richness and noise suppression in local routing.
\begin{table}[h]
\centering
\begin{tabular}{c c c}
\toprule
Local Radius ($r$) & Top-$k$ & bpc $\downarrow$ \\
\midrule
16 & 4  & 1.146 \\
16 & 8  & \textbf{1.141} \\
16 & 12 & 1.142 \\
32 & 8  & 1.143 \\
\bottomrule
\end{tabular}
\caption{Parameter analysis of the local similarity-aware routing module. Lower bpc indicates better performance.}
\label{tab:local_param}
\end{table}

\subsection{Performance under Varying Top-$K$ Expert Activations}
To systematically investigate the impact of varying the number of activated experts $K$ on model performance, we conducted a series of comparative experiments on Enwiki8 and Wikitext-103.The experimental results are summarized in Table~\ref{tab:moe_k_comparison}. 
Overall, as $K$ increases, most routing strategies show unstable or degraded performance, whereas MCF-MOE remains stable and consistently superior.

\begin{figure*}[t]
\centering
\begin{subfigure}[b]{0.48\textwidth}
    \centering
    \includegraphics[width=\textwidth]{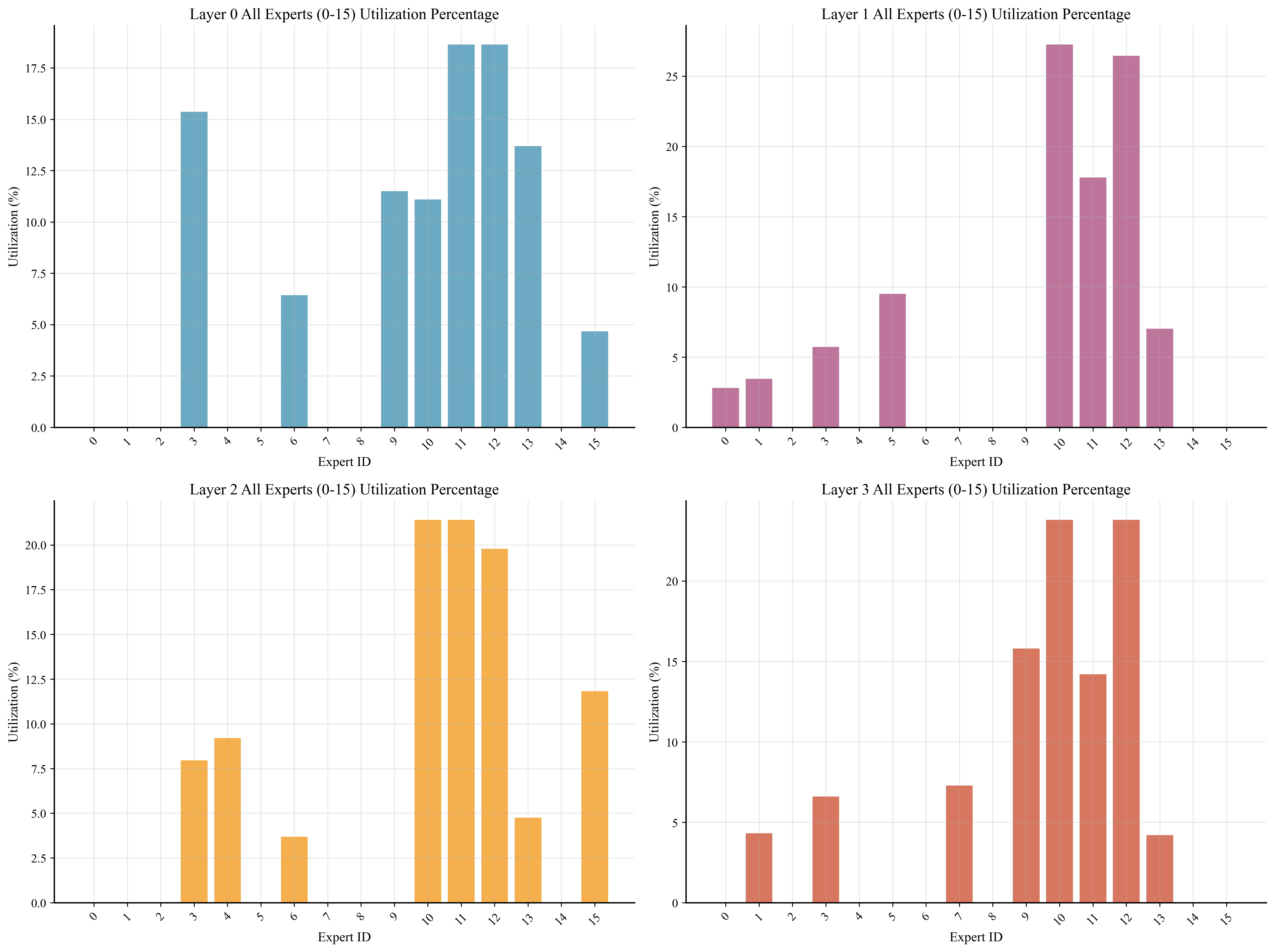}
    \caption{SMOE-Dropout}
    \label{fig:expert_usage_smoe}
\end{subfigure}
\hfill
\begin{subfigure}[b]{0.48\textwidth}
    \centering
    \includegraphics[width=\textwidth]{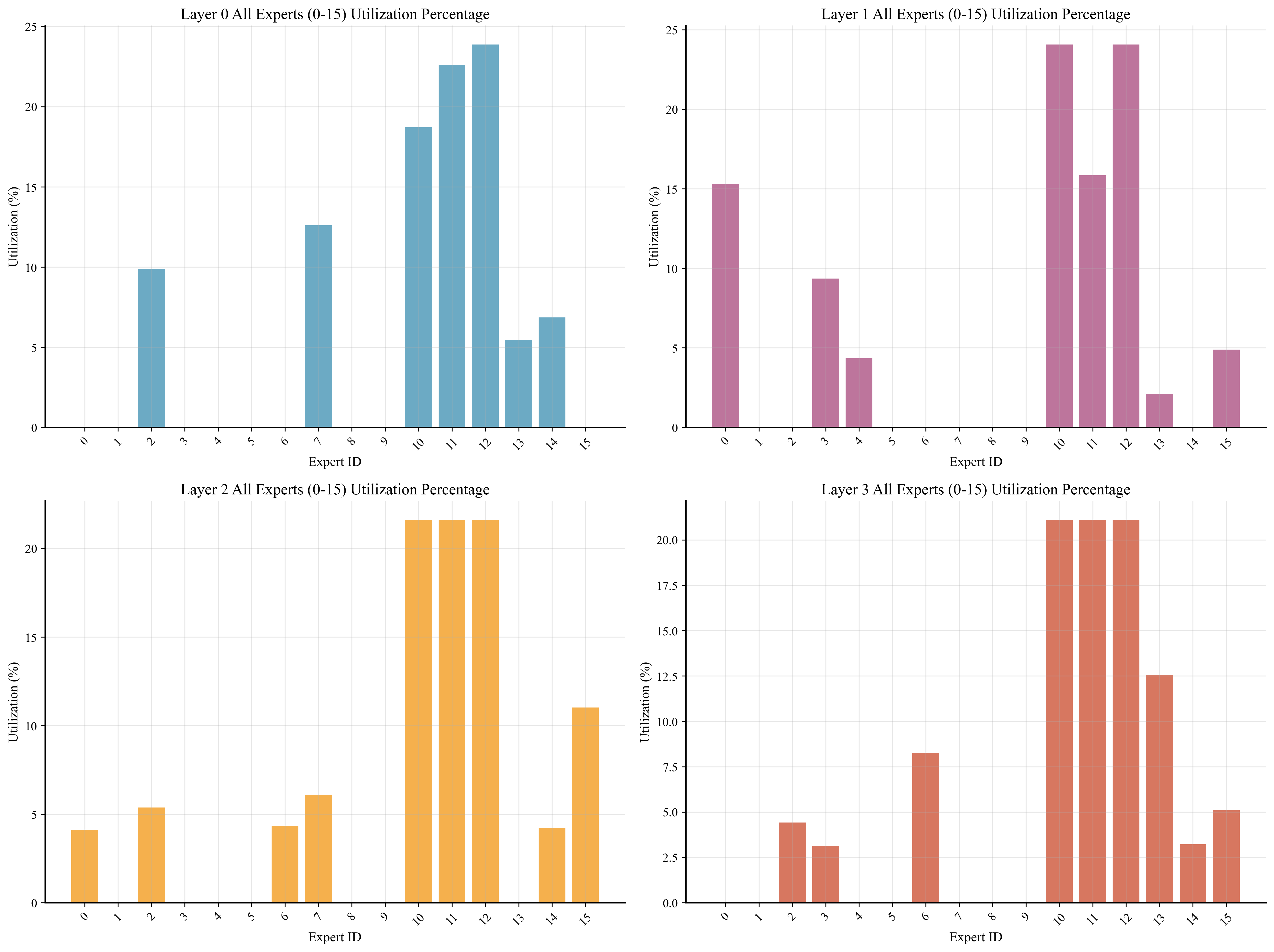}
    \caption{HyperRouter}
    \label{fig:expert_usage_hyper}
\end{subfigure}

\vskip\baselineskip 

\begin{subfigure}[b]{0.48\textwidth}
    \centering
    \includegraphics[width=\textwidth]{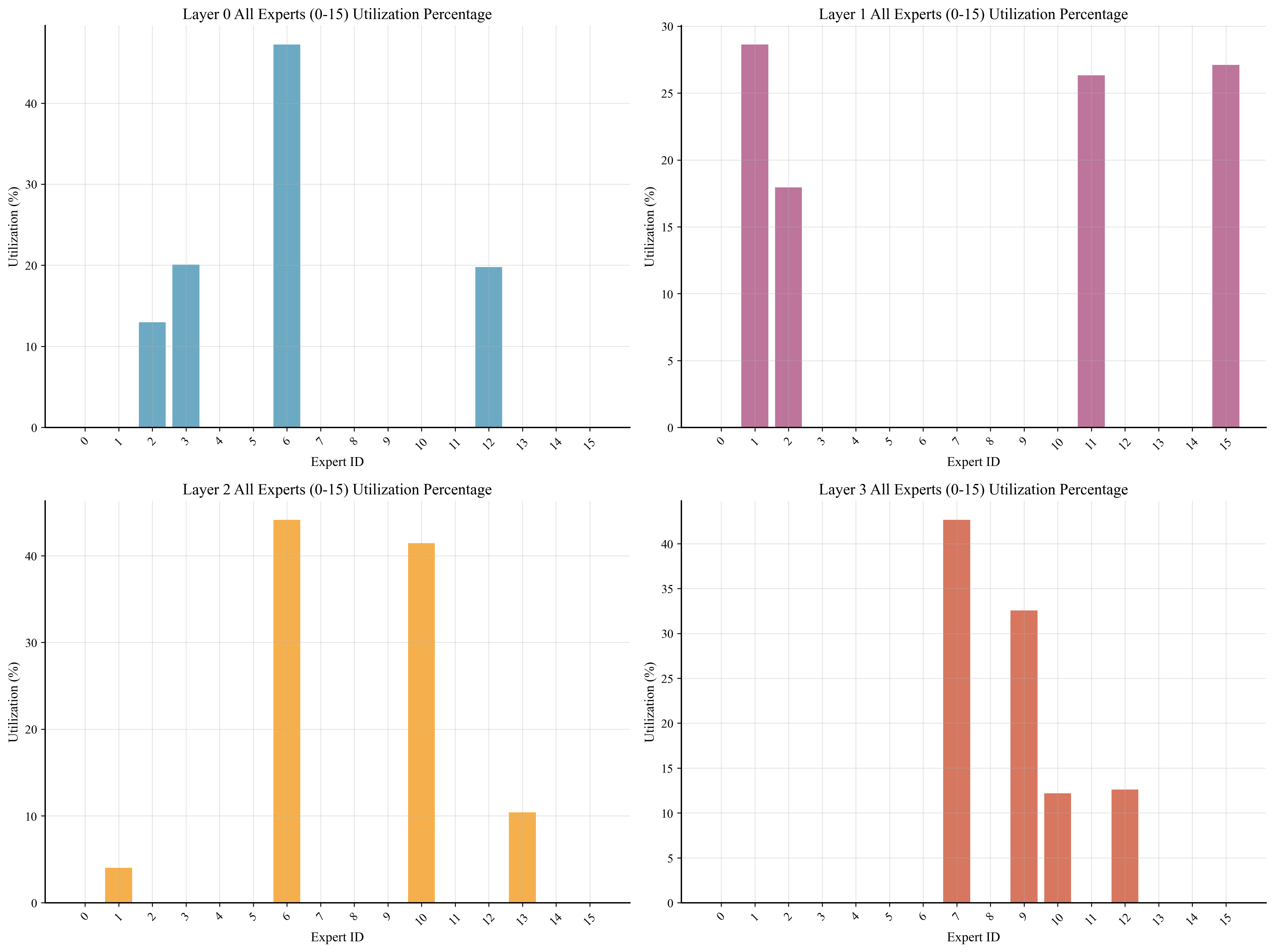}
    \caption{RMOE}
    \label{fig:expert_usage_rmoe}
\end{subfigure}
\hfill
\begin{subfigure}[b]{0.48\textwidth}
    \centering
    \includegraphics[width=\textwidth]{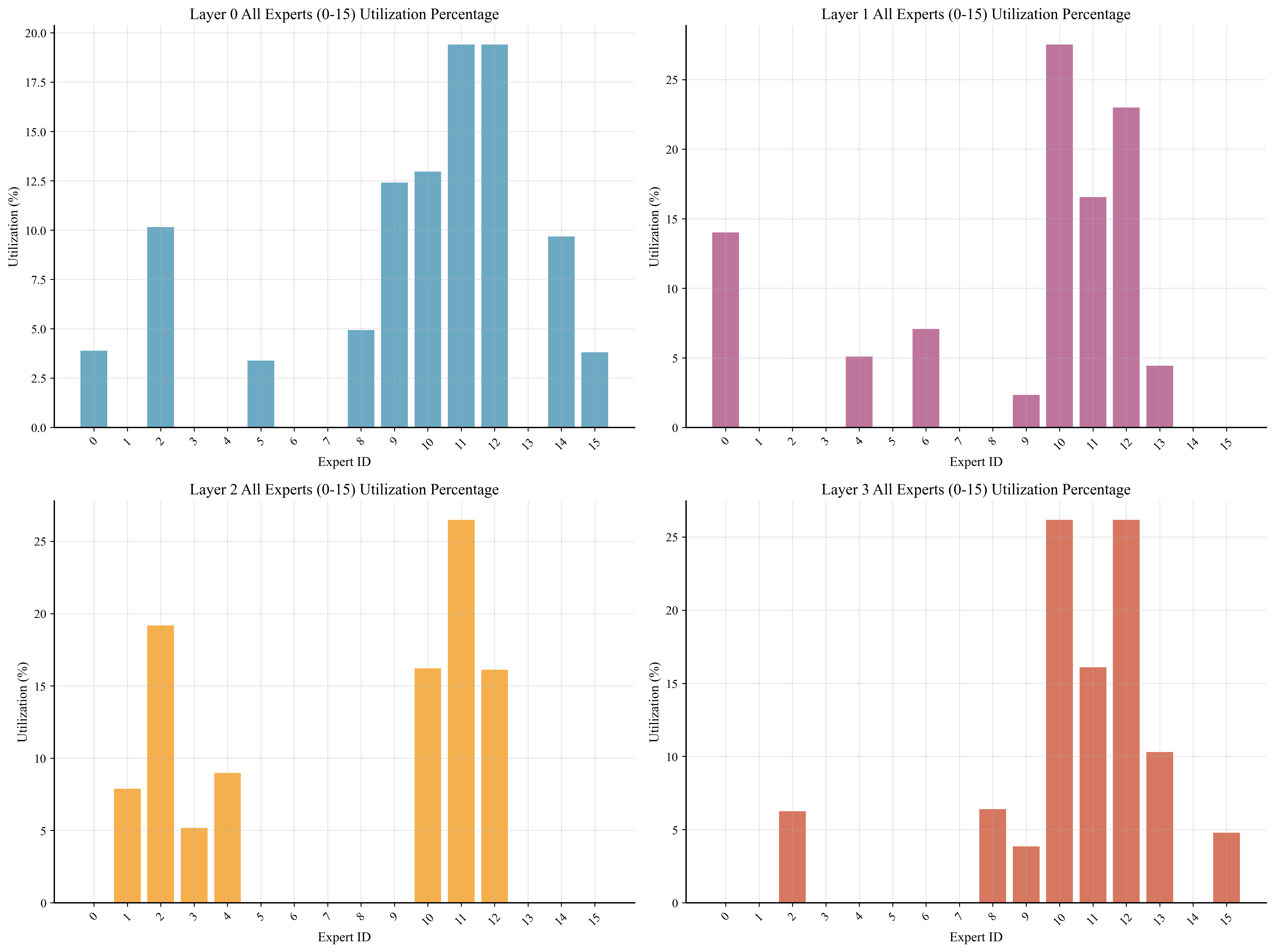}
    \caption{MCF-MOE}
    \label{fig:expert_usage_mcfmoe}
\end{subfigure}

\caption{Comparison of expert utilization across all layers in four routing methods.}
\label{fig:expert_utilization_4models}
\end{figure*}

On Enwiki8, conventional methods such as RMOE and SMOE-Drop are highly sensitive to changes in $K$, showing noticeable non-monotonic trends and significant degradation at larger $K$ values. In comparison, MCF-MOE maintains near-constant performance across different $K$ settings, consistently operating within an optimal range, indicating strong robustness to varying activation sizes. Similarly, on Wikitext-103, most baseline methods suffer from performance instability or decline as $K$ increases. In contrast, MCF-MOE consistently achieves lower perplexity across various $K$ configurations, outperforming other strategies and exhibiting superior generalization capabilities. This suggests that MCF-MOE is more effective in aligning local and global semantics in complex tasks such as language modeling, thereby enhancing expert consistency across tokens.

These findings indicate that MCF-MOE exhibits stronger adaptability and robustness under different Top-$K$ configurations. By alleviating the instability and semantic mismatch issues prevalent in existing routing strategies, MCF-MOE provides a more reliable solution for sparse expert models operating under dynamic computational constraints.

\section{Observation of Improved Routing Consistency through Multi-Level Context Integration}

\begin{figure}[h]
\centering
\includegraphics[width=0.45\textwidth]{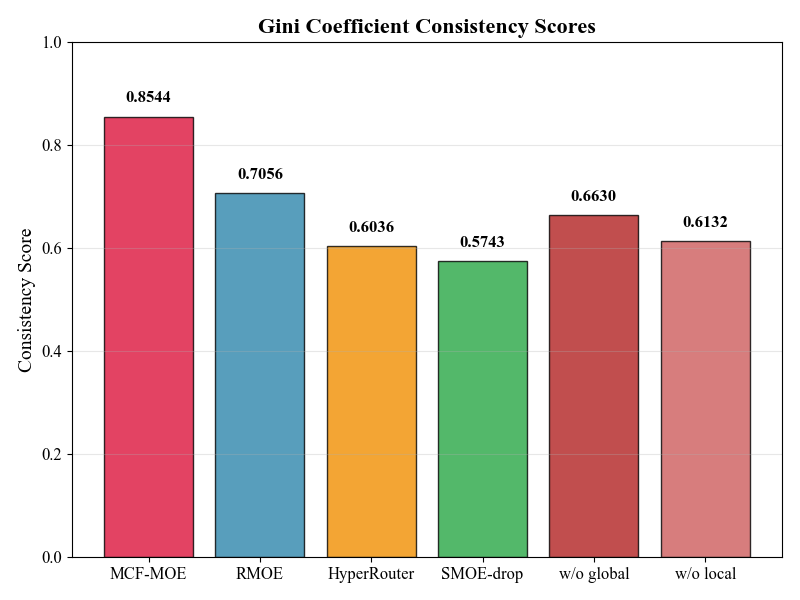} 
\caption{Gini consistency scores of MCF-MOE and baseline models, evaluates the stability and consistency of expert routing across different model variants. Higher is better.}
\label{Fig:gini}
\end{figure}

To evaluate expert selection consistency, we compare MCF-MOE with several baselines and ablated variants using the Gini coefficient, which measures alignment of routing decisions across semantically related tokens \cite{DBLP:journals/corr/abs-2411-07983}. Higher Gini values indicate more stable and efficient routing. As shown in Figure~\ref{Fig:gini}, MCF-MOE achieves the highest Gini score, significantly outperforming all baselines. In contrast, RMOE and HyperRouter—without multi-dimensional contextual modeling—show lower consistency, while SMOE-Dropout’s stochastic routing leads to the weakest alignment.

In addition, we evaluate two ablation variants of MCF-MOE: one without local context fusion and the other without global context fusion. Both variants exhibit noticeably lower Gini consistency compared to the full model, highlighting the complementary and mutually reinforcing roles of global and local context modeling in enhancing the consistency and stability of expert routing.

The superior performance of MCF-MOE highlights its ability to effectively fuse global abstractions with local structural cues, resulting in more consistent and reliable expert routing.

\section{Model scalability analysis }

\begin{figure}[h]
\centering
\includegraphics[width=0.45\textwidth]{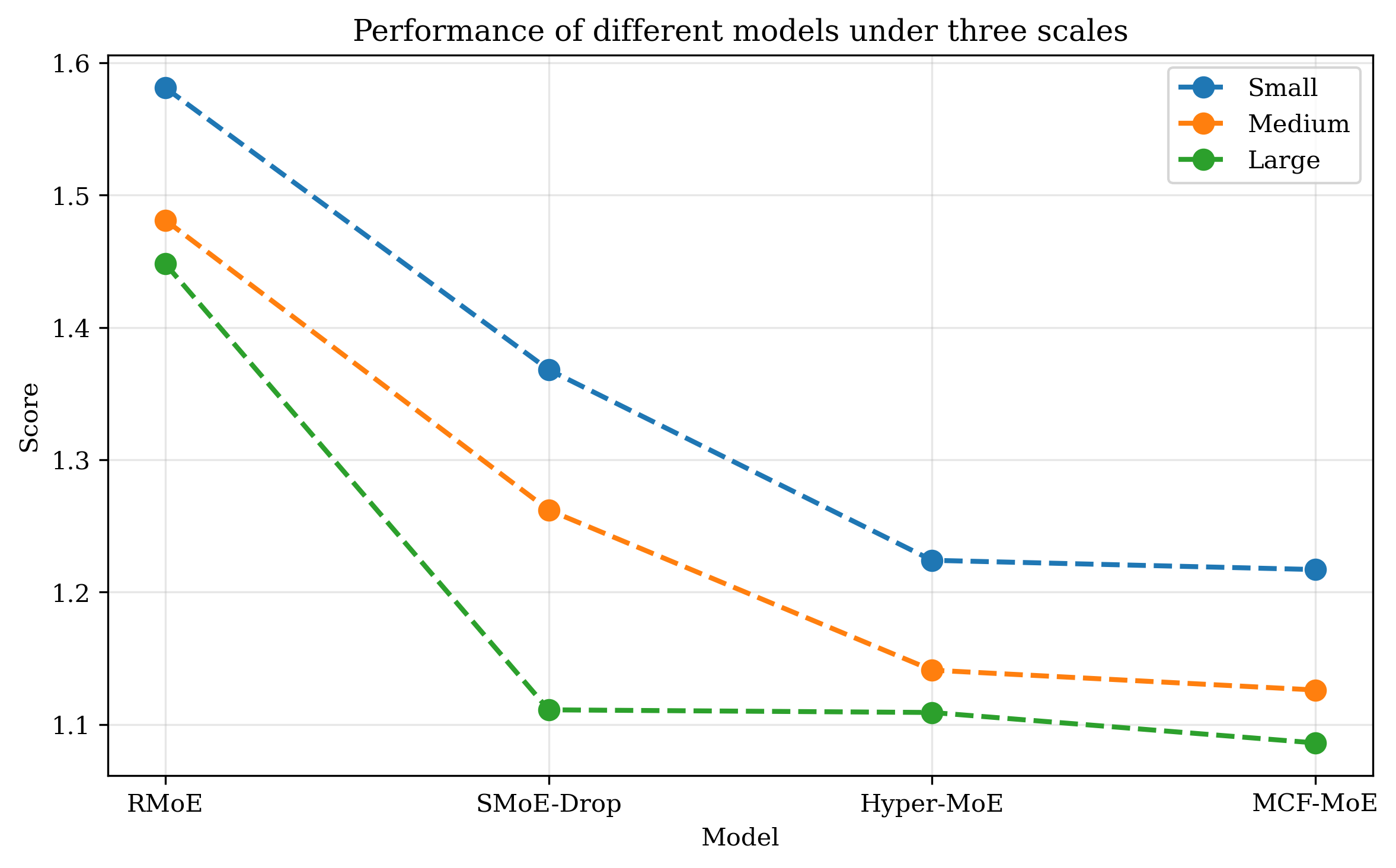} 
\caption{Performance of different gating mechanisms under three scales on the Enwiki8 dataset. Lower is better.}
\label{Fig:scale_up}
\end{figure}

To assess whether the benefits of MCF-MOE persist as model size increases, we conduct a parameter-scaling study following common practice in prior routing-focused MoE work (e.g., RMOE, HyperRouter, SMOE-Dropout), which typically evaluates routing behavior using small or mid-sized models. Constrained by compute, our main experiments use a modest configuration; however, to explicitly examine scalability, we further scale the model to 2× (Medium) and 3× (Large) of the original size and evaluate performance on Enwiki8 using bits-per-character (bpc; lower is better). The results are shown in Figure \ref{Fig:scale_up}.

Across all three scales, MCF-MOE consistently achieves the best performance:
1.217 → 1.126 → 1.086 (Small → Medium → Large), outperforming RMOE, SMOE-Drop, and HyperRouter at each level. Importantly, the relative ranking among methods remains unchanged as parameters grow, indicating that the improvements delivered by MCF-MOE arise from its routing-stability mechanism rather than from model size.

These results demonstrate that MCF-MOE not only provides gains in small-scale settings but also retains its advantage under model scaling, suggesting strong potential for future application to larger MoE architectures and full-scale LLM pretraining.


\section{Hierarchical Expert Utilization Analysis}

Figure~\ref{fig:expert_utilization_4models} presents the expert utilization across all layers for four models. SMOE-Dropout, HyperRouter, and MCF-MOE exhibit a relatively diverse expert activation pattern, with many different experts being engaged across layers, and a few (e.g., Expert 10–12) consistently demonstrating high usage. In contrast, RMOE activates significantly fewer experts per layer, resulting in a large number of experts that are rarely or never selected. This limited activation not only constrains the model’s ability to fully exploit the expert pool but also reflects inefficient resource utilization, which may hinder both generalization and robustness.

\section{Future Work}
In this work, we propose MCF-MOE, a sparse expert routing framework that fuses global cross-layer and local similarity-aware context to improve expert selection and collaboration. Experiments across multiple benchmarks validate its effectiveness and generalizability.

In the future, we will focus on addressing the limitations of MCF-MOE. One direction is to develop more robust cross-layer fusion strategies that can better handle semantic divergence in very deep decoder-only models, potentially through adaptive alignment mechanisms or layer-wise weighting schemes. Another direction is to reduce the sensitivity to hyperparameter choices by exploring automated tuning methods or self-adaptive fusion strategies, which could simplify training and improve stability across different model scales and tasks. In addition, investigating more efficient implementations to mitigate computational overhead and memory usage associated with multi-level context fusion would further enhance the practical applicability of MCF-MOE.

\end{document}